\documentclass[lettersize,journal]{IEEEtran}
\usepackage{amsmath,amsfonts}
\usepackage{algorithmic}
\usepackage{algorithm}
\usepackage{stfloats}
\usepackage{array}
\usepackage[caption=false,font=normalsize,labelfont=sf,textfont=sf]{subfig}
\usepackage{textcomp}
\usepackage{stfloats}
\usepackage{url}
\usepackage{verbatim}
\usepackage{graphicx}
\usepackage{cite}
\usepackage{booktabs} 
\usepackage{color}
\usepackage{subcaption}

\usepackage[inkscapelatex=false]{svg} 

\begin{document}

\title{Predictive Clustering of Vessel Behavior\\ Based on Hierarchical Trajectory Representation}

\author{Rui Zhang,~Hanyue Wu,~Zhenzhong Yin,~Zhu Xiao,~\IEEEmembership{Senior Member,~IEEE,}~Yong~Xiong,~and Kezhong Liu
\thanks{Manuscript received XX XX, 2023.}

\thanks{Rui Zhang, Hanyue Wu and Zhenzhong Yin are with the School of Computer Science and Artificial Intelligence, Wuhan University of Technology, Wuhan, 430000, China. E-mail:~\{zhangrui,~276817, yzzero\}@whut.edu.cn.
  }
\thanks{Zhu~Xiao and Yong~Xiong are with the College of Computer Science and Electronic Engineering, Hunan University, Changsha, 410082, China. E-mail:~\{zhxiao,~xiongyong\}@hnu.edu.cn.
  }
\thanks{Kezhong Liu is with the School of Navigation, Wuhan University of Technology, Wuhan, 430000, China. E-mail:~kzliu@whut.edu.cn.
  }

}

\markboth{Journal of \LaTeX\ Class Files,~Vol.~14, No.~8, August~2021}%
{Shell \MakeLowercase{\textit{et al.}}: A Sample Article Using IEEEtran.cls for IEEE Journals}


\maketitle
\begin{abstract}
Vessel trajectory clustering, which aims to find similar trajectory patterns, has been widely leveraged in overwater applications. Most traditional methods use predefined rules and thresholds to identify discrete vessel behaviors. They aim for high-quality clustering and conduct clustering on entire sequences, whether the original trajectory or its sub-trajectories, failing to represent their evolution. To resolve this problem, we propose a Predictive Clustering of Hierarchical Vessel Behavior (PC-HiV). PC-HiV first uses hierarchical representations to transform every trajectory into a behavioral sequence. Then, it predicts evolution at each timestamp of the sequence based on the representations. By applying predictive clustering and latent encoding, PC-HiV improves clustering and predictions simultaneously. Experiments on real AIS datasets demonstrate PC-HiV's superiority over existing methods, showcasing its effectiveness in capturing behavioral evolution discrepancies between vessel types (tramp vs. liner) and within emission control areas.  Results show that our method outperforms NN-Kmeans and Robust DAA by 3.9\% and 6.4\% of the purity score.
\end{abstract}

\begin{IEEEkeywords}
Trajectory clustering, Marine transportation, Data mining.
\end{IEEEkeywords}

\section{Introduction}
\IEEEPARstart{W}ITH the advancement of wireless positioning technology, numerous vessels are currently equipped with the Automatic Identification System (AIS). The AIS system is capable of collecting real-time information regarding vessel movements. Consequently, vessel trajectories are gathered as sequences of location points, speed, course, and navigation status throughout vessel voyages. Finding similar vessel trajectories is critical to overwater applications, e.g., anomaly detection \cite{ref3}, collision avoidance \cite{ref4}, and vessel behavior prediction \cite{ref24}. Clustering methods play a fundamental role in such applications.

 Trajectory clustering is mainly classified by two strategies according to hierarchies. The direct strategy is using similarity measures such as using Dynamic Time regular distance (DTW) \cite{ref9}, Hausdorff distance\cite{ref10,refhh1}, and Fr$\rm{\acute{e}}$chet distance\cite{ref12} to cluster the whole trajectory, in which it may be warped to match others. On the other hand, due to different sampling rates and unequal lengths of trajectories, some methods consider a fine-grained hierarchy, where they segment a trajectory into sub-trajectories, and cluster them instead. e.g., Gao \textit{et al.} identified the behavior patterns of sub-trajectories and used spectral clustering algorithms to define these patterns as the basis for vessel operation\cite{ref2}. Both of them aim for high-quality clustering and conduct clustering on whole sequences, whether the original trajectory or its sub-trajectories.
\begin{figure}[!t]
\centering
\includegraphics[width=3in]{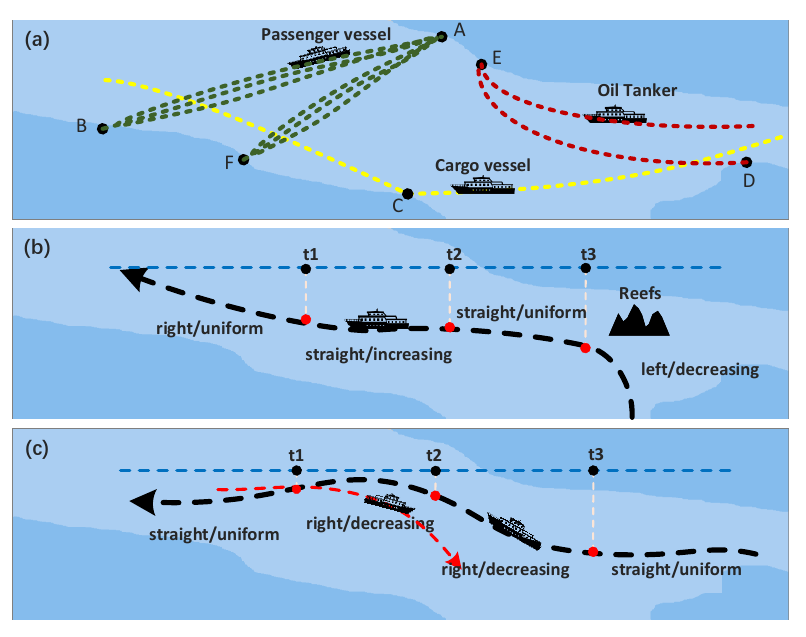}
\caption{Behavior sequence in different levels. (a)  shows the mooring behavior sequence of different vessels over the same time period. The green trajectory of the passenger vessel indicates that it has made multiple trips between specific ports. The yellow trajectory of the cargo vessel indicates that it only made one call at port C during this period; The red trajectory of the tanker indicates that it will moor in specific ports (E, D) with the berth of the tanker. The behavior evolution of the same type of vessels is similar based on the frequency of mooring, port preference, etc.. (b),(c) divide vessel trajectory into different stages according to the timestamp $t$ of the behavioral transition point. (b)~The vessel turns left, slows down to avoid hitting the reef, and then goes straight. (c)~Two vessels meet, according to the COLREGS\cite{ref5} rules, they will generally turn right to give way. As such, vessels have similar behavior sequences during the sailing process.}
\end{figure}

Apart from the original trajectory or its sub-trajectories, is it possible to find other representations to keep its pattern while making the clustering easier? To answer it, behavior sequence may be a good representation of trajectory. In this line, we define the behavior sequence as a serialization of phased changes in typical behaviors during the vessel's sailing process. On the one side, it overcomes the shortage of irregularity. On the other side, a behavior sequence can show the evolution over time, which provides a pattern description more clearly. Fig. 1 illustrates examples of trajectories and their behavior sequences. Considering all mooring behaviors happened over time of a trajectory, depending on frequency and positions, we can clarify whether it is of a passenger vessel, an oil tanker, or a cargo vessel (Fig. 1(a)). At the same time, a trajectory can be divided into sub-trajectories to show finer-grained evolution. For instance, when a vessel (black dashed, Fig. 1(b)) meets reefs (Fig. 1(b)) or another vessel (red dashed, Fig. 1(c)), it always turns and decreases speed. Several existing methods investigated using trajectories to recognize typical vessel behaviors. Nevertheless, they discuss no possible influence of evolution.

In this paper, we propose the Predictive Clustering of Hierarchical Vessel Behavior (PC-HiV) to extract behavior sequences from trajectories and find evolution. The PC-HiV first employs a hierarchical representation to transform a trajectory into three behavior sequences, including a position sequence, several sub-trajectory sequences, and a label sequence. Then, it clusters all sequences simultaneously to capture similar evolution. 
Specifically, the proposed clustering model (PC-HiV) strives to analyze the evolution of vessel behaviors and perform trajectory representations at different levels. In doing so, the proposed PC-HiV allows potential representations to be mapped to more interpretable discrete representations. By dividing the trajectory into multiple levels,  each level can reflect the evolution process of vessel behavior. As such, different levels of vessel trajectory features can be utilized for various tasks to achieve a balance between accuracy and efficiency. We conduct experiments on multiple levels and find that our hierarchical trajectory representation method provides more effective and meaningful clustering results. To the best of our knowledge, we are the first to construct trajectory features hierarchically and perform clustering on the behavioral evolution of vessels dynamically.

The contributions of this paper are summarized as follows. 

\begin{enumerate}
\item~We present hierarchical representations, leading to an enhanced synthesis of trajectories compared to traditional representations. At the level of the sub-trajectory sequence, the proposed PC-HiV retains more spatio-temporal information of the trajectory. At the level of the label sequence, the proposed method concentrates on specific behaviors, enabling the capture of long-term information.
\item~We demonstrate that similar behavior evolution among vessel trajectories can help clustering. Accordingly, we employ this insight to cluster vessel trajectories into different clusters over time, instead of clustering stable behaviors or entire behaviors. This leads to better effectiveness and performance.
\item{Experiments on real AIS dataset show the effectiveness of our method. The PC-HiV outperforms NN-Kmeans and Robust DAA by 3.9\% and 6.4\% in terms of the purity score, respectively.
~Our proposed clustering method, on the one hand, is applicable to represent the behavioral evolution differences between tramp shipping and liner shipping, on the other hand, can capture the behavioral evolution characteristics of vessels in emission control areas.}
\end{enumerate}

The remainder of this paper is organized as follows. Section~\ref{sec:Related Work} discusses related works for vessel trajectory clustering and predictive clustering. 
Section~\ref{sec:Preliminaries} presents the preliminary and definitions.~After that, Section~\ref{sec:Methodolology} details the design of the proposed PC-HiV. Section~\ref{sec:Experiments} presents the experiments compared with baselines. Finally, Section~\ref{sec:conclusion} concludes the paper.

\section{Related Work}\label{sec:Related Work}
\subsection{Vessel Trajectory Clustering}
This section discusses two main strategies for vessel trajectory clustering, i.e., direct strategies and fine-grained strategies. To specify, the direct strategies mostly focus on the overall distribution characteristics of the vessel trajectory. The fine-grained strategies consider both the behaviors and evolution of vessel trajectory.

For the direct strategies, vessel trajectory clustering captures the spatio-temporal features of the overall trajectory. Existing studies on vessel trajectory clustering are based on trajectory compression methods or trajectory feature extraction. Various trajectory feature extraction methods can exhibit significant impact on vessel trajectory clustering. For example, Jiang Q \textit{et al.} compressed the trajectory with the Douglas-Peucker (DP) algorithm and measured the overall trajectory distance with the Hausdorff method~\cite{refhh1}. X. Xu \textit{et al.} compressed the vessel trajectory using the DP algorithm and then applied the DBSCAN clustering multiple times to reduce the number of points while keeping the reference route's characteristics intact~\cite{ref11.3}. J. Yang \textit{et al.} used the DP algorithm to compress vessel trajectories and delete a large number of redundant data points~\cite{ref11.4}. However, compressing and clustering the trajectory may lead to loss of information and degradation of the vessel's continuous behavior change process.

Xiao \textit{et al.} proposed a lattice-based DBSCAN algorithm to extract the vessel’s channel\cite{ref6}. Considering that DBSCAN is less capable of obtaining the expected results by clustering trajectories of different densities,  the HDBSCAN, namely, a hierarchical density-based method was proposed to cluster vessel trajectories. For instance, Wang \textit{et al.} combined Dynamic Time Regularization (DTW) and HDBSCAN to identify the main routes and speed profiles~\cite{ref9}. L. Eljabu \textit{et al.} suggested a spatial clustering approach for maritime traffic that relies solely on dynamic time warping (DTW) similarities and does not use traditional clustering methods~\cite{ref11.5}. Wang L \textit{et al.} proposed a clustering method for vessel trajectories via using HDBSCAN and Hausdorff distance measurement~\cite{ref10}. Tang \textit{et al.} proposed the vessel trajectory clustering method FOLFST, which is hierarchically clustered to obtain trajectory clusters of the same channel~\cite{ref11}. Cao \textit{et al.} used Fr$\rm{\acute{e}}$chet distance to realize adaptive vessel trajectory clustering to improve the performance of trajectory clustering~\cite{ref12}. Although promising, the above-mentioned methods only clustered the shapes or distribution features of vessel trajectories. In addition, those methods usually suffer from high computational costs and sensitivity to noise and data with non-uniform sampling rates. 

Recently, several learning models such as RNNs and CNNs have strong feature representation capabilities and achieved considerable success in trajectory similarity computation~\cite{ref13}. In this line, H. Duan \textit{et al.} employed one-hot encoding to represent the attributes of a vessel's trajectory, such as LAT, LON, SOG, COG, WID, LEN, and DRA, helping to better understand the potential meaning of the spatiotemporal information of the trajectory. On the downside, this approach led to an increase in data dimensionality~\cite{ref11.1}.
M Liang\textit{et al.} combines vessel motion features and trajectory geometry features simultaneously. The motion features are defined to extract vessel features, while they tried to map the trajectory image into the low-dimensional feature embeddings~\cite{ref11.2}. Y. Zhang \textit{et al.} proposed a PDGCN method for clustering of target domains, combining clustering and coding for training in an end-to-end manner~\cite{refw}. G. Shi \textit{et al.} represented the samples as category embedding and superclass embedding, which are generated by using k-means clustering on existing base classes, and the embedding representation enhanced by the fusion of two embedding vectors~\cite{refw3}. Zhang \textit{et al.} presented a robust auto-encoder model with an attention mechanism to learn the low-dimensional representations of noisy vessel trajectories~\cite{ref14}. Liang M \textit{et al.} proposed a convolutional auto-encoder neural network, remapping the trajectories into informative trajectory image matrices, thereby improving trajectory clustering performance~\cite{ref15}.

For the fine-grained strategies, most existing research is based on predefined vessel behaviors or focused on a single behavior. Those methods are not able to detect unforeseen behaviors and require the assistance of domain specialists for deﬁning rules and thresholds~\cite{ref23}. For example, Ma \textit{et al.} uses a spectral clustering algorithm to extract the motion pattern of the vessel~\cite{ref16}. Such a method focused on a single behavior, regarding the movement from Port A to Port B as a type of behavior. Huang \textit{et al. } applied a semantic analysis method to explore potential mobility patterns by combining semantic transformation and topic model~\cite{ref17}. The vessel behaviors obtained with this method are discrete and fail to serialize the evolution. Besides, researchers considered typical behaviors of vessels. For example, Gao \textit{et al.} identified the behaviors of sub-trajectories and used spectral clustering algorithms to define these behaviors as the basis for vessel operation~\cite{ref2}. Jia \textit{et al.} proposed an auto-encoder method based on a two-stream LSTM network that extracts the motion primitives of vessels from encounter data~\cite{ref18}. Note that the clustering algorithm based on complete trajectories ignores the similar behavior evolution of specific sub-trajectory segments, and results in low efficiency and low detection accuracy; the clustering algorithm based on trajectory points ignores the spatial correlation between trajectory points. 

\subsection{Predictive Clustering}
Predictive clustering is widely used in time-series clustering, such as the predictive clustering tree methods introduced by Džeroski \textit{et al.}. This method used traditional distance metrics for sequences and identified clusters that minimize the distance variance of each leaf, and then assigned a label to each cluster based on the centroid\cite{ref28}. Predictive clustering tree is built on the basis of a decision tree whose leaves can represent a cluster. In recent years, researchers have proposed a variety of predictive clustering methods. Harbi \textit{et al.} added supervised learning predictions to unsupervised k-means and divided the same type of label objects into a cluster. They combined the simulated annealing with the improved K-means algorithm to classify objects with the same label into the same group~ \cite{ref30}. Van \textit{et al.} proposed a constraint-based clustering method that exploits background knowledge to construct clusterings~\cite{ref31}. Lemaire \textit{et al.} provided interpretability for clustering results by combining the predictive power of supervised learning with unsupervised clustering \cite{ref32}. 

Changhee Lee’s work \cite{ref19} is a deep learning method for clustering time series data, AC-TPC is performed by learning discrete representations that best describe the distribution of future results based on the loss function, encouraging each cluster to have homogeneous results. It enables the mutual improvement between cluster assignment tasks and label prediction tasks.  Similar to the work done by J. Gou \textit{et al.}, focus more attention on important areas and less attention on irrelevant areas\cite{refw2}. Motivated by those, we apply the idea of predictive clustering to vessel behavior clustering, where the homogeneity of clusters is measured by the similarity of behavior labels.

\section{Preliminaries}\label{sec:Preliminaries}
Original vessel trajectories are obtained through AIS, generally require preprocessing prior to subsequent analysis and exploration\cite{ref19.1,ref19.2}. To capture the behavioral patterns of vessels, preprocessed trajectories are transformed into the hierarchical structure of vessel trajectories, including position sequence, sub-trajectory sequence and label sequence. 

\textbf {Definition 1 (Position Sequence). }A position sequence $T$ is a sequence of position points $p_i$: $T = \{p_1 \rightarrow  p_2 \rightarrow ... \rightarrow p_n\} \in \mathbb{R}^{n \times d} $, where $n$ is the number of positions, and $p_j \in \mathbb{R}^{d_1}$ contains a $d_1$-dimensional feature vector at position $j$, including coordinates, speed, course, timestamp, etc. $T$ is ordered by timestamps.

\textbf {Definition 2 (Sub-Trajectory Sequence). }A sub-trajectory sequence $T^{\prime}$ is a sequence of sub-trajectory $T_i$: $T^{\prime}= \{T_1 \rightarrow T_2 \rightarrow ... \rightarrow T_m\} \in \mathbb{R} ^{n \times d_i}$.

Here, each $T_i$ is a subsequence of a time series $T$, which is a continuous subset of points from $T$ with a length of $k$, where $1 \leq i_1 \leq i_k \leq n$. A position sequence $T$ can be represented by m mutually exclusive sub-trajectories, i.e.,  $ \forall 1\leq i,j\leq m, T_i \cap T_j = \emptyset$ and they construct a new sequence $T^{\prime}= \{T_1 \rightarrow T_2 \rightarrow ... \rightarrow T_m\} \in \mathbb{R} ^{n \times d_i}$. The difference between $T$ and $T^{\prime}$ is slices, where $T$ can be seen as a special case of $T^{\prime}$ when $k = n, \forall i$.

\textbf {Definition 3 (Label Sequence). }A label sequence $T^{\ast}$ of status $p^{\ast}_{i}: T^{\ast} = \{p^{\ast}_1 \rightarrow p^{\ast}_2 \rightarrow ... \rightarrow p^{\ast}_ m\} \in \mathbb{R}^{m \times d_2} $. For each $p^{\ast}_ i$, through a mapping function $M: \mathbb{T}\rightarrow \mathbb{P}^{\ast}$, where $\mathbb{T}$ is the set of $T_{i}$ and $\mathbb{P}^{\ast}$ is the set of $p^{\ast}$. 

$M$ maps sub-trajectories to a certain $p\ast$, which includes position information and some vessel status information, e.g.decreasing speed, mooring. Denote it as the label point $p^{\ast}_i \in \mathbb{R}^{d_2}$. 

For clarifying the notations,  we provide Table~\uppercase\expandafter{\romannumeral 1} that presents the deceptions of main variables used in this paper.

\begin{table}[!t]
    \centering
    \caption{Summary of notations} 
    \begin{tabular}{cp{6cm}}
    \toprule
        Notation & Description \\ 
        \midrule
        $p_{i}$ & position point that contains lon, lat, speed, course, timestamp, etc\\
        $T$ & sequence of position points $p_{i}$ \\
        $T_{i}$ & sub-trajectory of $T$\\
        $\mathbb{T}$ & set of $T_{i}$\\
        $y_i$ & behavior label of $T_i$\\
        $Y$ & behavior label sequence composed of $y_i$\\
        $\mathbb{P}$ & set of $p^{\ast}_i$\\
        $T^{\prime}$ & sequence consist of $T_{i}$ \\
        $p^{\ast}_{i}$ & a status of the whole $T_{i}$ including position information and some vessel status information.\\
        $T^{\ast}$ & sequence of status points $p^{\ast}_{i}$\\
        $y_i$ & behavior label corresponding to $T_{i}$\\
        $B$ & behavior label set\\
        $y_i^{\ast}$ & label corresponding to $p^{\ast}_{i}$\\
        $Y^{\ast}$ & sequence composed of $y_i^{\ast}$\\
        $N$ & total number of trajectories\\
        $C l s_{i}$ & all samples in the $i$-th category\\
        $c_{i}$ & all the real sample in the $i$-th category\\
        $C l s$ & all the clusters\\
        $\mathbb{C}$ & true category\\
        $K$ & number of clusters\\
        \bottomrule
    \end{tabular}
\end{table}

Using hierarchical representations, we can represent a position sequence as some sub-trajectories, and a label sequence, from the bottom to the top. Fig. 2 shows an example of them. By such representations, there is a clear transformation from a lower to a higher hierarchy, which provides more information to help clustering.
\begin{figure}[!t]
\centering
\includegraphics[width=2.5in]{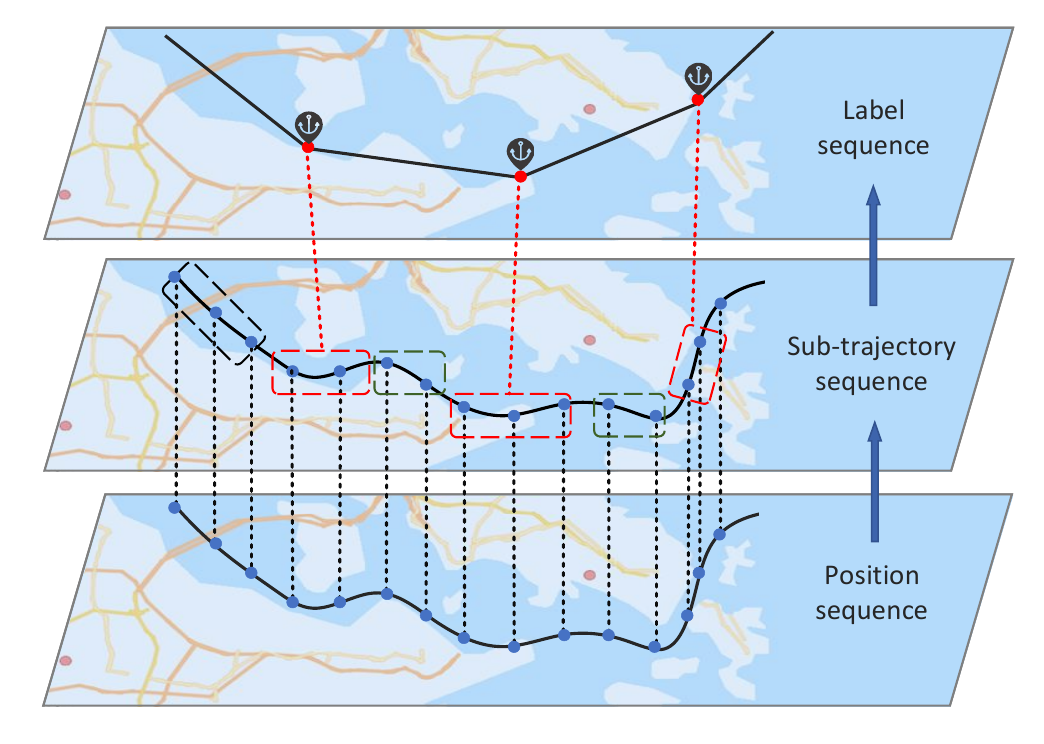}
\caption{Hierarchical vessel trajectory structure. One can observe a mapping relationship from the continuous position sequence to the sub-trajectory sequence, as well as a mapping relationship from the sub-trajectory sequence to the label sequence. Each segment in trajectory represents a behavior. In addition, the specific behaviors in the sub-trajectory are concerned and the sub-trajectory is mapped to a label point to form a label sequence.}
\label{fig_2}
\end{figure}

With the above representations, we can now use predictive clustering to simultaneously solve clustering and prediction tasks. 

\textbf {Problem Statement:} Given a position sequence $T$, our task is to find $K$ best clusters $Cls = \{Cls_1, \ldots, Cls_K\}$ and then summarize evolution. Each group's trajectories can be represented by a centroid based on some similarity measure, and the group assigned to each vessel’s behavior should be updated as time progresses.

We will discuss evolution results in Sec.\uppercase\expandafter{\romannumeral5} and show hierarchical representations provide a better explanation of trajectories.

\section{Methodology}\label{sec:Methodolology}
In this work, we propose the Predictive Clustering of Hierarchical Vessel Behavior (PC-HiV) to represent trajectories in hierarchies and predicatively cluster behaviors. To specify, the proposed PC-HiV first leverages the hierarchical representation to gain three behavior sequences, as defined in Sec.~\ref{sec:Preliminaries}.  Along with the capturing of behavior sequences, labels are generated to reflect behaviors and used in the predictive clustering. After that, the PC-HiV encoders behavior sequences, and utilizes the encoded representations to cluster and predict behaviors. Fig. 3 shows the overall framework of our proposed method.
\begin{figure*}[h]
\centering
\includegraphics[width=6.5in,keepaspectratio]{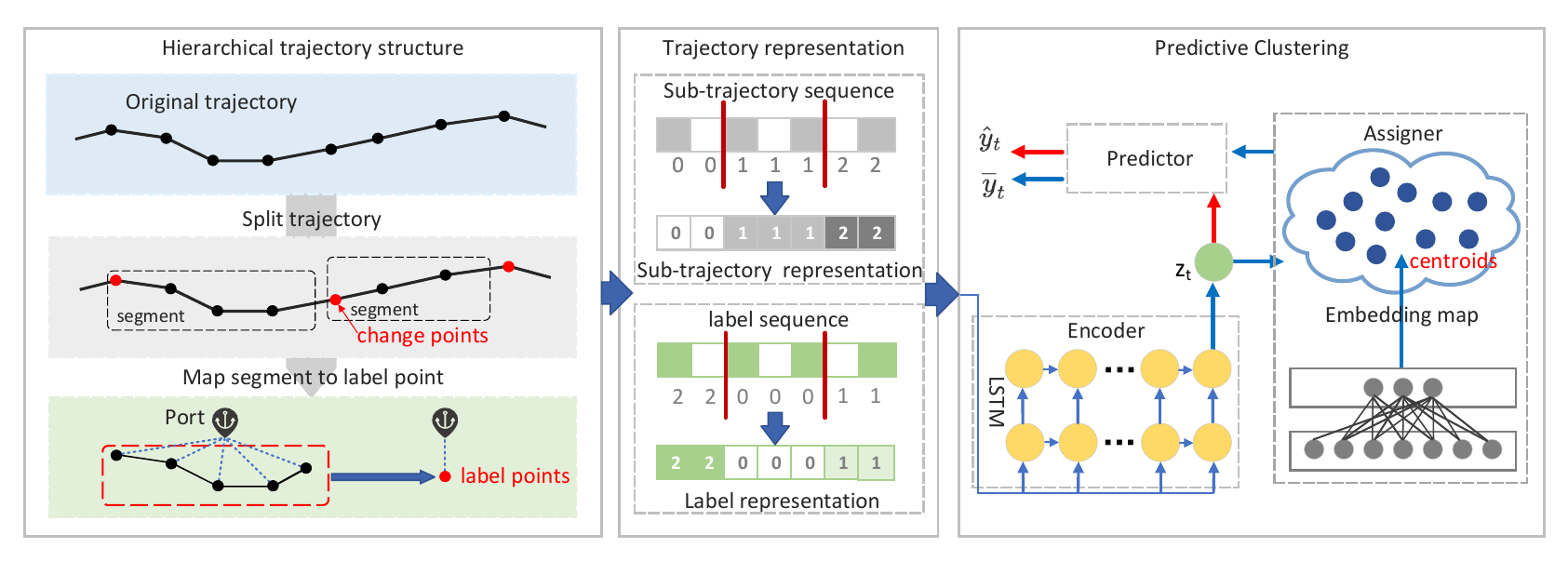}
\caption{PC-HiV model framework. It consists of a vessel trajectory hierarchy representation module, label generation module, encoder, predictor, and assigner. }
\label{fig3}
\end{figure*}

\subsection{Hierarchical trajectory representation}
According to Sec.~\ref{sec:Preliminaries}, given a position trajectory, it automatically becomes a position sequence $T$, which is the first hierarchy. Next, the PC-HiV recursively searches change points of $T$, and uses these points to gather a sequence of sub-trajectories $T^{\prime}$ as the second hierarchy. To that end, we pre-segment the trajectory according to a fixed length $u$, and obtain the pre-change point $pre$. We define the change points as those $p_j$  that have the largest spatial-temporal distances between their former points and latter points. Note that their scores are larger than threshold $\delta$, indicating that two sub-trajectories $T_i$s are mostly not similar. Accordingly, each $T_i$ can reflect a behavior. Then, the PC-HiV distributes behavior labels for each $T_i$ according to a predefined behavior set $B$. We consider features of $T_i$, such as speed changes or turns, then find the closest behavior $y_i \in B$.

For each sub-trajectory segment, we divide its speed into decreasing, increasing, uniform speed, and stopping. For the direction, it goes left, right, and straight. Speed and direction can jointly determine ten types of vessel behaviors. Considering speed and direction separately, we calculate the speed difference between the two positions of the sub-trajectory segment, if 80\% of the speed difference of the adjacent points is in the same direction, both are positive (negative), then it is regarded as increasing (decreasing). Otherwise, we calculate the variance of the speed of the sub-trajectory segment, if it is less than the threshold and has small change in speed (the uniform speed case). The special case is that the speed is less than 10 nautical miles/hour as a stop. If it is greater than the threshold, we calculate the speed difference between the endpoint and the starting point. Additionally, if it is positive (negative), then responding to accelerated (deceleration). Considering the direction, we rely on the difference of the vessel's heading to the ground. If it is greater than the threshold, it turns right. If it is greater than the threshold, we define straight line moving, otherwise, it is a left turn.

\begin{algorithm}
\caption{Represent sub-trajectory sequence}\label{alg:1.1}
\begin{algorithmic}
\STATE \textbf{Input:} position sequence:$T$; number of segment $S$; behavior set: $B$; number of nearest segments to calculate: $\lambda$; spatial-temporal distance threshold:$\delta$; pre-change points $pre$ $\leftarrow$ $\{p_u,p_{u*2}...,p_{u*(n/u-1)}\}$
\STATE \textbf{Output:} sub-trajectory representation
\STATE change point set:$C$ $\leftarrow$ $\emptyset$
\FOR{$p_i$ $\in$ $pre$} 
    \STATE $T_{left}\leftarrow \{(p_i-\lambda*u),...,p_i\}$
    \STATE $T_{right}\leftarrow \{p_i,...,(p_i+\lambda*u)\}$
    \STATE $score_{p_i}\leftarrow \mathrm{dis}(T_{left},T_{right})$
\ENDFOR
\STATE $C$ = $\{ p_i| score_{p_i}>\delta\}$
\STATE $T^{\prime} \leftarrow {\rm segment}(T,C)$
\FOR{$T_i \in T^{\prime}$}
    \STATE behavior label:$y_i$ $\leftarrow$ Matching the behavior of $T_i$ in set $B$
\ENDFOR
\STATE sub-trajectory representation: $T^{\prime}$(sub-trajectory sequence) and $Y$(sequence of $y_i$) 
\end{algorithmic}
\end{algorithm}

Following that, the label sequence $T^*$ is generated from $T ^{\prime}$ as the last hierarchy. As mentioned above, a $p _i^* \in T^*$ marks vessel status information which renders $T^*$ and provides a coarse-grained behavior evolution. In this paper, we utilize unique metrics from trajectory data, the so-called dwell time, to help further understand travel behavior~\cite{refh1}. We can select a specific behavior $\mathit{\Upsilon}$, such as vessel mooring behavior, or other significant maneuvering behavior. We take the segments with behavior $\mathit{\Upsilon}$ from the sub-trajectory sequence. Then we can map each sub-trajectory segment to a point representing its status. In doing so, choosing different behaviors can find different evolution characteristics of vessel behavior. For example, we use mooring status to construct $p^*_i$ as it shows the purpose of that vessel. The $T^*$ of a passenger’s vessel may only have a few fixed mooring points, since it makes round trips between ports, while the $T^*$ of an oil tanker may have plenty of mooring points because it stays for a long time. Specifically, the proposed PC-HiV chooses all $T_i$s that have mooring $p_{i_j} $s and matches them with port positions. When categorizing berths and determining their coordinates, we accomplish this by two steps: \textit{i}) count the number of vessels that have moored at a particular berth; \textit{ii}) select the one with the highest proportion as the category of the berth. For each $T_i$, the first $p_{i_j}$ will be the label (mooring) point $p^*_ i$, if it satisfies that the distance between it and a port is less than a given threshold. If $T_i$ matches many ports, then $p^*_ i$  denotes the point that has the minimum distance to its port. 

After generating $T^*$, we use port information in this hierarchy as it relates to label points. The mooring preference label $y^*_ i$ describes a port $P_i$ that contains most types of vessel trajectories. Note that the categorization does not mean that port $i$ only has such vessels. Following that, we gain all behavior sequences and their representations, which can be used for predictive clustering. Algorithm 1 and Algorithm 2 show the process of representing a trajectory in hierarchies.

\begin{algorithm}[h]
\caption{Represent label sequence}\label{alg:1.2}
\begin{algorithmic}
\STATE \textbf{Input:} sub-trajectory: $T^{\prime}$ and $y_i$; behavior: $\mathit{\Upsilon}$; label set: $P$; all trajectory $\mathcal{T}$;
\STATE \textbf{Output:} label representation
\STATE $\hat T$$\leftarrow$$\{T_i|T_i \in T^{\prime} \And y_i=\mathit{\Upsilon}\}$
\FOR{$T_i \in \hat T$}
    \FOR{$P_j \in P$}
        \STATE $p_{behavior_j} \leftarrow \{p_{i_j}|{\rm F}(p_{i_j},P_j)<\sigma\}$ \COMMENT{${\rm F}$ is to judge the degree of match between point $p_{i_j}$ and label $P_j$}
        \STATE $p_{i_j}^* \leftarrow p_{behavior_j}[0]$
        \STATE $p_{i_j}^{\prime} \leftarrow p_{behavior_j}[mid]$
    \ENDFOR
    \STATE $contain\_label \leftarrow \{P_j|p_{behavior_j} \neq \emptyset\}$
    \IF{$|contain\_label|==0$}
        \STATE $\hat T$ delete $T_i$
    \ELSIF{$|contain\_label| \ge 2$}
        \STATE $p^*_i \leftarrow \underset{p^{\prime}_{i_j}}{\text{arg min }} {\rm F}(p^{\prime}_{i_j},{P}_j)$ 
    \ELSE
        \STATE $p^*_i \leftarrow p^*_{i_j}$, where $ P_j \in contain\_label$
    \ENDIF
\ENDFOR
\STATE $T^* \leftarrow $ $\{p_i^*\}_{i=1}^{i=|\hat T|}$
\FOR {$P_j \in P$}
    \STATE $y_{behavior_j} \leftarrow \underset{k}{\arg \max }\left|\mathrm{C}_k(\mathcal{T})\right|$
\ENDFOR
\FOR {$p_i^* \in T^*$}
    \STATE $y_i^* \leftarrow y_{behavior_j}$, where $p_i^*$ matches $P_j$
\ENDFOR
\STATE label representation: $T^{\ast}$(label sequence) and $Y^*$(sequence of $y_i^*$) 
\end{algorithmic}
\end{algorithm}
\subsection{Predictive Clustering}
The predictive clustering starts with encoding. Given a trajectory and its $T^{\prime}$, each $p_j$ has corresponding timestamp. According to \cite{ref51}, we utilize an LSTM network to encode each $p_j$ to its latent variable $z_j$, where $j$ can be seen as the $j$-th timestamp of $z \in \mathbb R^{n \times d_3}$. Here $d_3$ denotes the latent dimension. For $T^{\ast}$, as several some $p_j$ may not have label points $p_j^{\ast} $, we use the last point that has a label point to represent all points between two $p_j^{\ast}s \in T^{\ast}$. Besides, not all vessels park at a port simultaneously, all $z$s of  $T^{\ast}$s begin with the same $j$. That is to say, for each vessel, their absolute timestamps will be discarded. Instead, we shift all timestamps to the same beginning and use relative timestamps for the vessel to keep temporal information. 

Next, we use predictive clustering~\cite{ref19} to cluster each $z_j$ and predict behaviors. It has an assigner and a predictor. An assigner \textit{f} predicts the
probability $\mathcal{P}\left(k=1,2, \ldots, K \mid z_j\right)$ of the $k$-th cluster. $z_j$ will be assigned to the cluster with the highest probability. Then, a predictor \textit{g} predicts behaviors from $z_t$ and its corresponding cluster center $E(z_t)$. For $T^{\prime}$, the predictor calculates the probability $\mathcal{P}\left(y_i \in Y \mid z_j\right)$of $z_j$ and $\mathcal{P}\left(y_i \mid E\left(z_j\right)\right)$, respectively. The loss function of $z_j$ is calculated as follows. 
\begin{equation}
\begin{split}
\mathcal{L}_1=\mathbb{E}_{T^{\prime} \in \mathcal{T}}\left[\sum_{j=1}^n \sum_k l\left(y^{(j)}, g\left(z_j\right)\right) \cdot f_k\left(z_j\right)\right]
\end{split}
\end{equation}

\begin{equation}
\mathcal{L}_2=\mathbb{E}_{T^{\prime} \in \mathcal{T}}\left[\sum_{j=1}^n \sum_k l\left(y^{(j)}, g\left(E\left(z_j\right)\right)\right) \cdot f_k\left(E\left(z_j\right)\right)\right]
\end{equation}
where $\mathcal T$ is the set of all trajectories, $y^{(j)}$ is the behavior label of $z_j$ and $E(z_j)$, and $l\left(y^{(j)}, g(x)\right)=-\sum_i y_i^{(j)} \log g_i(x)$. For $T^{\ast}$, everything is the same except the label is $Y^{\ast}$.

The model achieves a balanced prediction of clusters by minimizing the KL divergence. The predicted label distribution for each group should be homogeneous. Beyond that, it is to be effectively represented by the group's centroid. Therefore, each determined group can possess two characteristics. On the one hand, at each evolution stage, vessels of the same type exhibit similar behavior. On the other hand, the type assigned to each vessel's behavior is updated as time progresses. To achieve this, the model transforms the predictive clustering problem into learning a discrete representation, namely learning labels that can reflect the process of the entire vessel behavior evolving up to the current evolution stage. As shown in Fig. 3, the blue arrows represent the process of estimating the probability distribution ${\bar{y}}_t$ from the representations of cluster centroids. The red arrows represent the process of directly estimating behavioral labels from the latent representation $z_t$, with the estimated probability distribution being ${\hat{y}}_t$. The model minimizes the KL divergence as the evaluation metric for the similarity between two probability distributions, ${\bar{y}}_t$ and ${\hat{y}}_t$.

Compared to traditional trajectory clustering methods, PC-HiV has two features. First, by hierarchical representations, it can cluster and analyze behaviors in different granularities, which provides richer information. Then, since predictive clustering learns representations of each trajectory, the PC-HiV supports updating clusters (representations) with time. This is especially beneficial to vessels, because they may have new trajectories and behaviors at once. Therefore, the proposed PC-HiV can reflect the evolution of vessel behaviors.

\section{Experiments}\label{sec:Experiments}
\subsection{Dataset}
We use AIS \cite{ref1} data from the Port of Ningbo-Zhoushan, China from March 1 to March 31, 2015, and the Port of Hainan, China from October 1 to October 31, 2018, to evaluate the effectiveness of the proposed method. We perform data quality check, trajectory smoothing, and slicing operations in sequence to clean the trajectory data and prevent trajectories from repeating along specific routes. AIS data has various features of vessels that can be used by PC-HiV. In the experiments, we mainly use MMSI, latitude, longitude, SoG, CoG, timestamp, and vessel type. We filter all trajectories with $n \geq 3, 000$ and remove sparse trajectories of which the interval between two points is greater than half hour. The size of final trajectories is $\sim 1M/day$. The total water area is $\sim$70,000 $km^2$.

\subsection{Settings}
We use a single-layer LSTM~\cite{ref51} with a hidden size = 150 as the encoder. Both assigner and predictor are a two-layer MLP with a hidden size = 50. We use Adam~\cite{ref52} as  optimizer, with initial learning rate = 0.001, $\beta 1$ = 0.9, $\beta2$ = 0.999. All modules apply dropout \cite{ref53} with a probability of 0.7. Our models are implemented by TensorFlow. For sub-trajectory sequence experiments, there are 10 behaviors in total, i.e., the combination of one category of speed status (accelerated, decelerated, uniform, stopped) and one turning status (left, right, straight). In particular, we do not associate the stop state with the turning status. For label sequence experiments, all points will be classified as one out of three ports with mooring behavior preference.

We adopt Purity, ARI~\cite{ref54} and NMI~\cite{ref22}, to evaluate the partitioning effectiveness of the proposed PC-HIV. Purity ranges from (0, 1), describing the degree to which clustered samples belong to the same category. ARI and NMI are used to compare the similarity between clustering results and true class labels. ARI ranges from (-1, 1) and NMI ranges from (0, 1). They follow the principle of 'the higher, the better'. The three metrics are defined in the following.
\begin{equation}
\mathrm{Purity}=(Cls, \mathbb{C})=\frac{1}{N} \sum_k \max _j\left|C l s_k \cap c_j\right|
\end{equation}

    \begin{equation}
        \mathrm{ARI} = \frac{\sum_{i}\sum_{j} \binom{n_{ij}}{2} - [\frac{\sum_{i} \binom{a_{i}}{2} \sum_{j} \binom{b_{j}}{2}}{\binom{N}{2}}]}{\frac{1}{2} [\sum_{i} \binom{a_{i}}{2}+\sum_{j} \binom{b_{j}}{2}] - [\frac{\sum_{i} \binom{a_{i}}{2} \sum_{j} \binom{b_{j}}{2}}{\binom{N}{2}}]}
    \end{equation}
    where $\mathrm{n}_{ij}$ is the element in the $i$-th row and $j$-th column of the confusion matrix, indicating samples belonging to true class $i$ and assigned to clustering result $j$. $a_{i}$ and $b_{j}$ are the number of samples in true class $i$ and the number of samples in clustering result $j$, respectively. $\binom{n_{ij}}{2}$ denotes the Binomial coefficient of the element in the $i$-th row and $j$-th column, representing the logarithm of samples correctly assigned in the same cluster.  $\frac{\sum_{i} \binom{a{i}}{2} \sum{j} \binom{b_{j}}{2}}{\binom{N}{2}}$ is an adjustment term.

    \begin{equation}
        \mathrm{N M I}(C l s, \mathbb{C})=\frac{2 \mathrm{M I}(C l s, \mathbb{C})}{\mathrm{H}((C l s)+\mathrm{H}(\mathbb{C})}
    \end{equation}
where $\mathrm{H}$ calculates the entropy of the random variable, $\mathrm{MI}$ calculates the amount of information shared between two random variables.

The baselinse include KM-DTW \cite{ref55}, NN-k-means \cite{ref19}, DCN-pro \cite{ref20}, SOM-VAE \cite{ref21}, Robust DAA\cite{ref14}, and AC-TPC \cite{ref19}.

\subsection{Clustering Performance}
Table \uppercase\expandafter{\romannumeral2} shows the performance comparison of sub-trajectory sequences. The proposed PC-HiV reaches the best performance on all three metrics. Since KM-DTW, DCN-pro, and SOM-VAE have no labels, their ARI is close to 0, which means their clusters are the same as randomly assigning each trajectory to any cluster. The reason behind this is that all those methods consider none or very limited features of evolution. KM-DTW and DCN-pro only use trajectory positions and even use no timestamp to show change with time, though DCN-pro takes much effort on mapping positions to low-dimension representations. As such, although DCN-pro has a better purity score and NMI than KM-DTW, indicating that it has better clustering results, they only show trajectories that have similar positions. SOM-VAE leverages the Markov model to capture spatial-temporal features. Nevertheless, the Markov model has a strong assumption that every point is only related to the previous one \cite{ref21}. It is not suitable for trajectories that have changes in a certain period. Besides, this shortage is particularly heavier for finding evolution.
\begin{table}[!tp]
\caption{Evaluation of sub-trajectory sequence clustering results}
\centering
\setlength{\tabcolsep}{4mm}
\begin{tabular}{cccc}
\hline  
Method & Purity & NMI & ARI\\
\hline  
KM-DTW & 0.514 & 0.117 & 0.061\\
NN\_Kmeans(z) & 0.715 & 0.429 & 0.279\\
NN\_Kmeans(y) & 0.793 & 0.532 & 0.426\\
DCN-pro & 0.725 & 0.198 & 0.043\\
SOM-VAE & 0.581 & 0.239 & 0.012\\
\textbf{PC-HiV[s]} & \textbf{0.832} & \textbf{0.640} & \textbf{0.590}\\
\hline 
\end{tabular}
\end{table}

Our PC-HiV and NN-k-means are both labeled methods. Also, NN-k-means uses latent representation. According to the clustering target is z or y, NN-k-means have two versions, i.e., NN-k-means (z) and NN-k-means (y). Labeling is essential for trajectory clustering because labels will correlate clusters to behaviors, even if those behaviors are in a period. Compared to unlabeled methods, labeled methods improve performance on every metric. It is not surprising that NN-k-means (y) is better than NN-k-means (z), because NN-k-means (y) clusters trajectories directly by labels, which makes clusters more homogeneous than NN-k-means (z). However, even if NN-k-means (y) uses strong supervision information, the proposed PC-HiV achieves better performance. There are two main reasons. First, PC-HiV updates the encoder, assigner, and predictor simultaneously, making them instruct each other when training. Apart from this, the cluster center is updated in a real-time manner, which is more flexible than NN-k-means. Then, PC-HiV uses an assigner to replace traditional k-means, overcoming the shortage that k-means is purely defined by its input and has no other information.

Table \uppercase\expandafter{\romannumeral3} depicts the performance comparison on label sequences. KM-DTW does not perform well in terms of Purity and NMI. AC-TPC  uses predictive clustering while extracting only position trajectories. Hence, it can be equivalence as our method without hierarchical representations. Overall, the proposed PC-HiV achieves a better purity score and ARI than AC-TPC, which shows the effectiveness of hierarchical representations. It is noteworthy that PC-HiV has dropped $\sim$ 0.1 NMI to AC-TPC. This is because,  in this experiment, we focus on the mooring behavior and only use label sequences with it, but vessels definitely have other common behaviors. Since AC-TPC uses all position trajectories, in the clustering process it may consider other behaviors thereby increasing NMI. A similar phenomenon turns out in grid-based Robust DAA. Although Robust DAA’s idea is different from sequence-based methods, it reaches high NMI as long as it uses position trajectories containing more behaviors. In summary, we show that PC-HiV learns good trajectory representations and the hierarchical representations can provide other information to improve clustering results.
\begin{table}[!t]
\caption{Evaluation of label sequence clustering results}
\centering
\setlength{\tabcolsep}{3mm}
\begin{tabular}{cccc}
\hline  
Method & Purity & NMI & ARI\\
\hline  
KM-DTW & 0.5261 & 0.1881 & 0.1585\\
AC-TPC\cite{ref19} & 0.6000 & 0.4736 & 0.1932\\
Robust DAA\cite{ref14} & 0.6775 & \textbf{0.4858} & 0.3169\\
\textbf{PC-HiV[l]} & \textbf{0.7418 } & 0.3720 & \textbf{0.3259}\\
\hline 
\end{tabular}
\end{table}

\subsection{Ablation Study} Fig. 4 illustrates the purity score, NMI, and ARI with the varying $K$ in different methods. PC-HiV obtains the best performance and is stable for all $K$ and all metrics. This means that PC-HiV can tolerate different situations of trajectories. Accordingly, compared to baselines, PC-HiV requires no pre-defined $K$.
\begin{figure}[h]
    \centering
    \subfloat[Purity]{
    \includegraphics[width=1.5in]{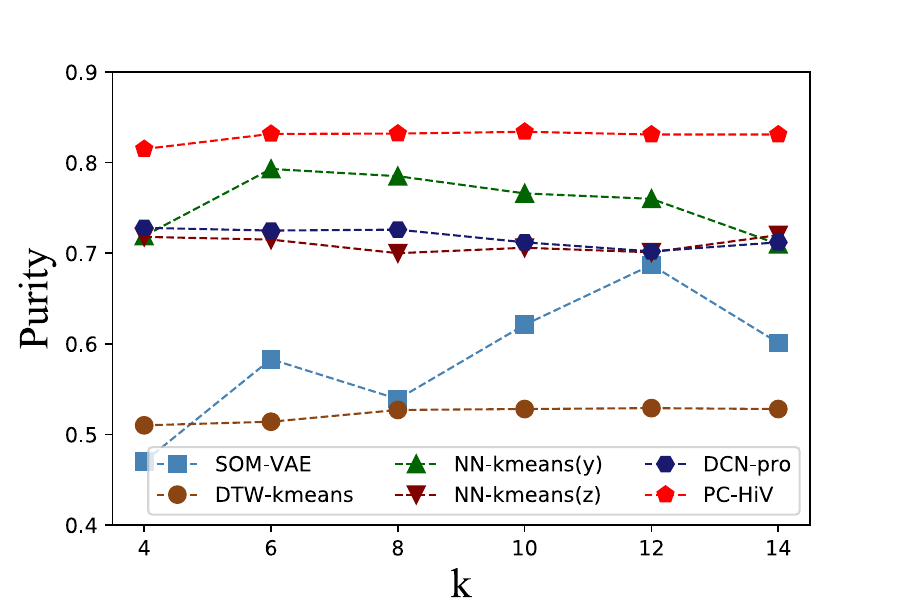}}
    \quad
    \subfloat[NMI]{
    \includegraphics[width=1.5in]{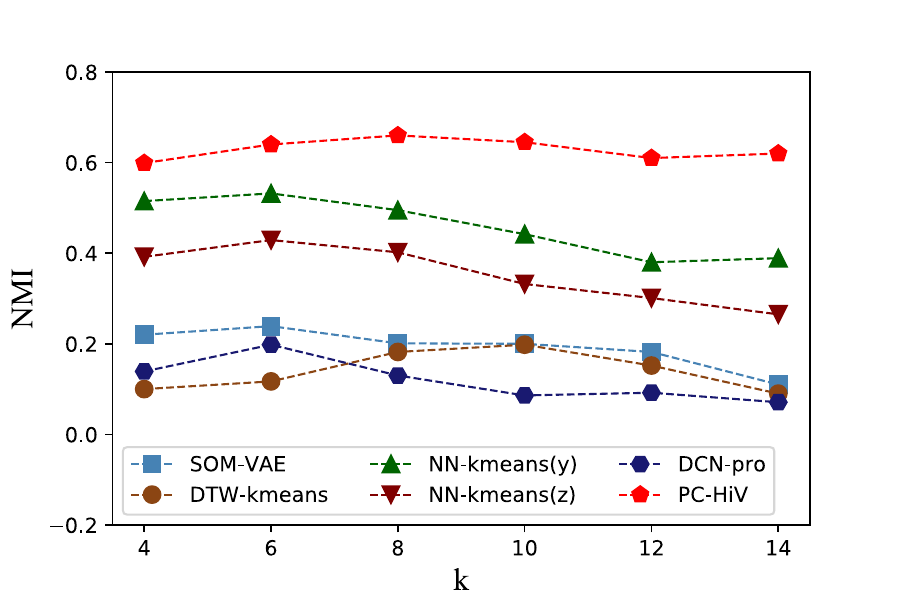}}
    \quad
    \subfloat[ARI]{
    \includegraphics[width=1.5in]{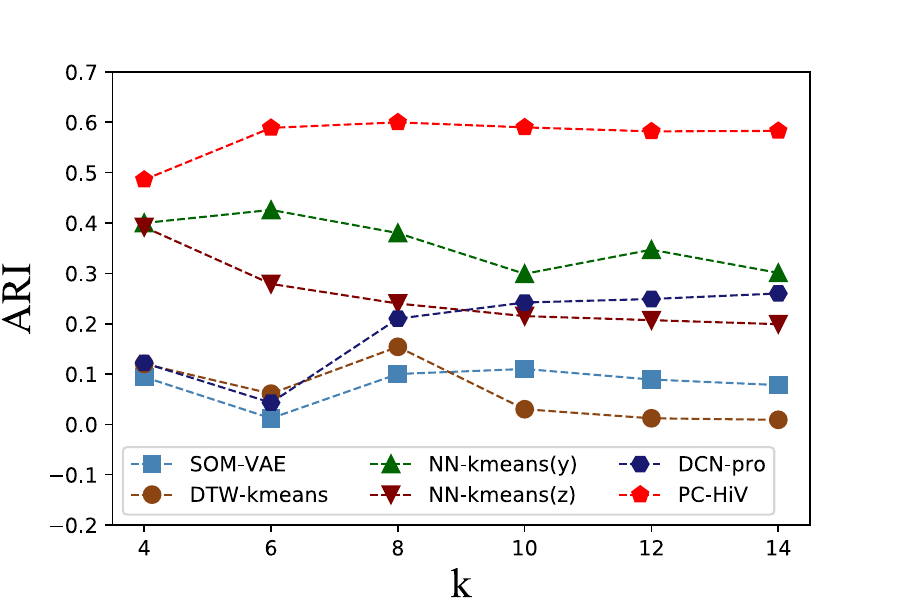}}\\
    \caption{Evaluation of vessel sub-trajectory sequence clustering results with different \textit{K}}
    \label{}
\end{figure}

\begin{figure}[h]
\centering
\includegraphics[width=0.5\textwidth]{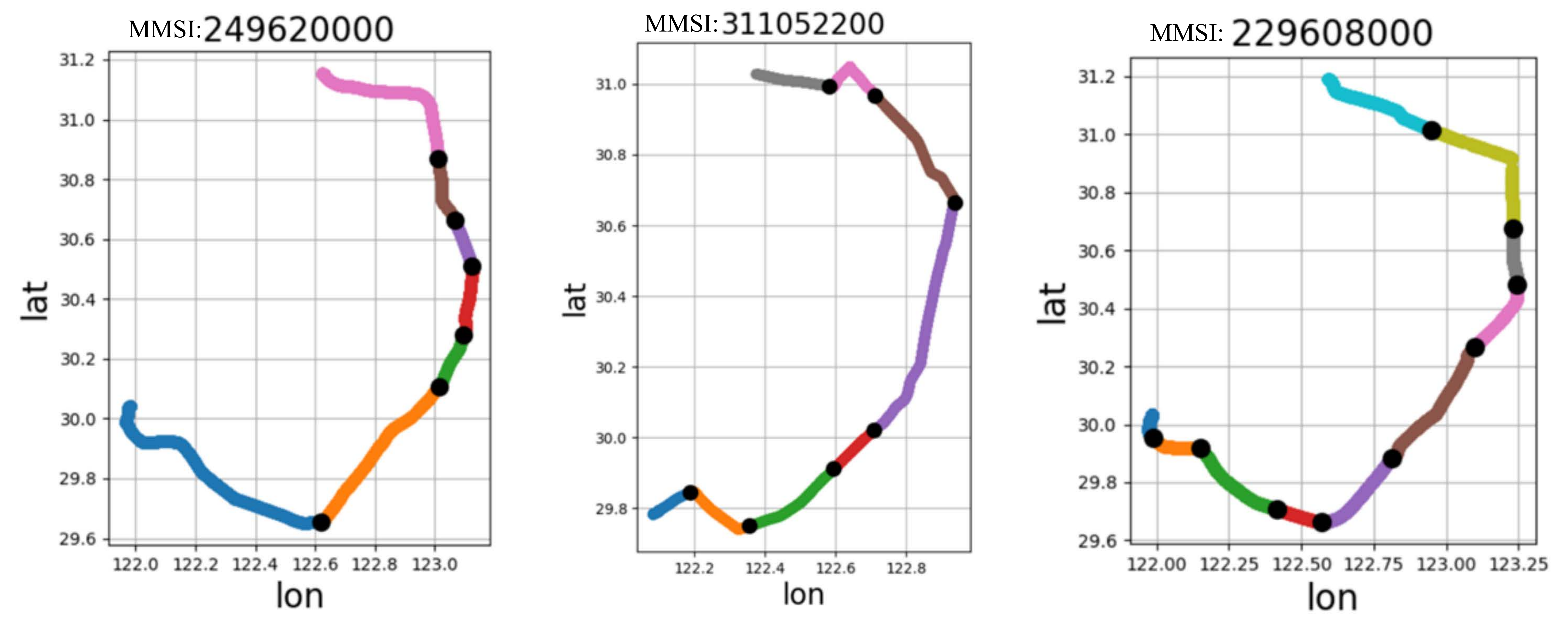}
\caption{Phased evolution of vessels in the same cluster}
\label{fig_3}
\end{figure}

Further, we investigate how the different representations influence PC-HiV’s performance. Specifically, we generate sub-trajectories by different methods, including speed threshold, direction threshold, AutoPlait ~\cite{ref56} and ClaSP ~\cite{ref57}. Table \uppercase\expandafter{\romannumeral5} shows the performance of these methods. Since AutoPlait and ClaSP separate a trajectory based on fine-grained features, they have better results than threshold-only methods. In other words, the fine-grained representation is more effective than coarse-grained representation, thereby enabling PC-HiV to reach good performance. On the other hand, PC-HiV outperforms baselines, showing that hierarchical representation can synthesize trajectory features better.
\begin{table}[!t]
\caption{Evaluation of vessel behavior clustering results for different sub-trajectory segment method}
\centering
\setlength{\tabcolsep}{2mm}
\begin{tabular}{cccc}
\hline  
Method & Purity & NMI & ARI\\
\hline  
speed threshold+PC-HiV& 0.185 & 0.145 & 0.073\\
Heading threshold+PC-HiV & 0.201& 0.363 & 0.272\\
AutoPlait+PC-HiV & 0.496 & 0.452 & 0.347\\
ClaSP+PC-HiV & 0.794 & 0.601 & 0.526\\
\textbf{our method+PC-HiV} & \textbf{0.831 } & \textbf{0.640} & \textbf{0.589}\\
\hline 
\end{tabular}
\end{table}
\subsection{Case study}
\subsubsection{Case of Evolution in the same cluster}
Fig. 5 shows an example of our found evolution of three different trajectories (The mmsi of these three vessels are: 249620000, 311052200, 229608000) within a cluster. Although the duration of each behavior of the three trajectories is different, PC-HiV successfully recognized them as one kind of evolution, because their behaviors are similar and happened in the same order.
\subsubsection{Case of Tramp Shipping and Liner Shipping}
In the study on Tramp Shipping and Liner Shipping clustering, we analyze 200 trajectories, each containing 2,858 data points. These trajectories are divided into two categories: container vessels (representing liner shipping) and oil tankers (representing tramp shipping). We compile separate lists of mooring locations for each vessel type. By comparing the latitude and longitude of each data point to these locations, we assign a value of +1 if the point is near the container vessel berth. Otherwise, we assign -1. Our model exhibits satisfactory clustering performance for two vessel types, with NMI: 0.6164, ARI: 0.6519, and PURITY: 0.9037. The clustering results are visualized in Fig.~6, with (a) displaying the actual trajectory types, and (b) illustrating the clustering outcomes. "Container" vessels are represented by Cluster 1 and Type 1, while "tanker" vessels are represented by Cluster 0 and Type 0.
\begin{figure*}
    \centering
    \subfloat[True classification]{
    \includegraphics[width=2.4in]{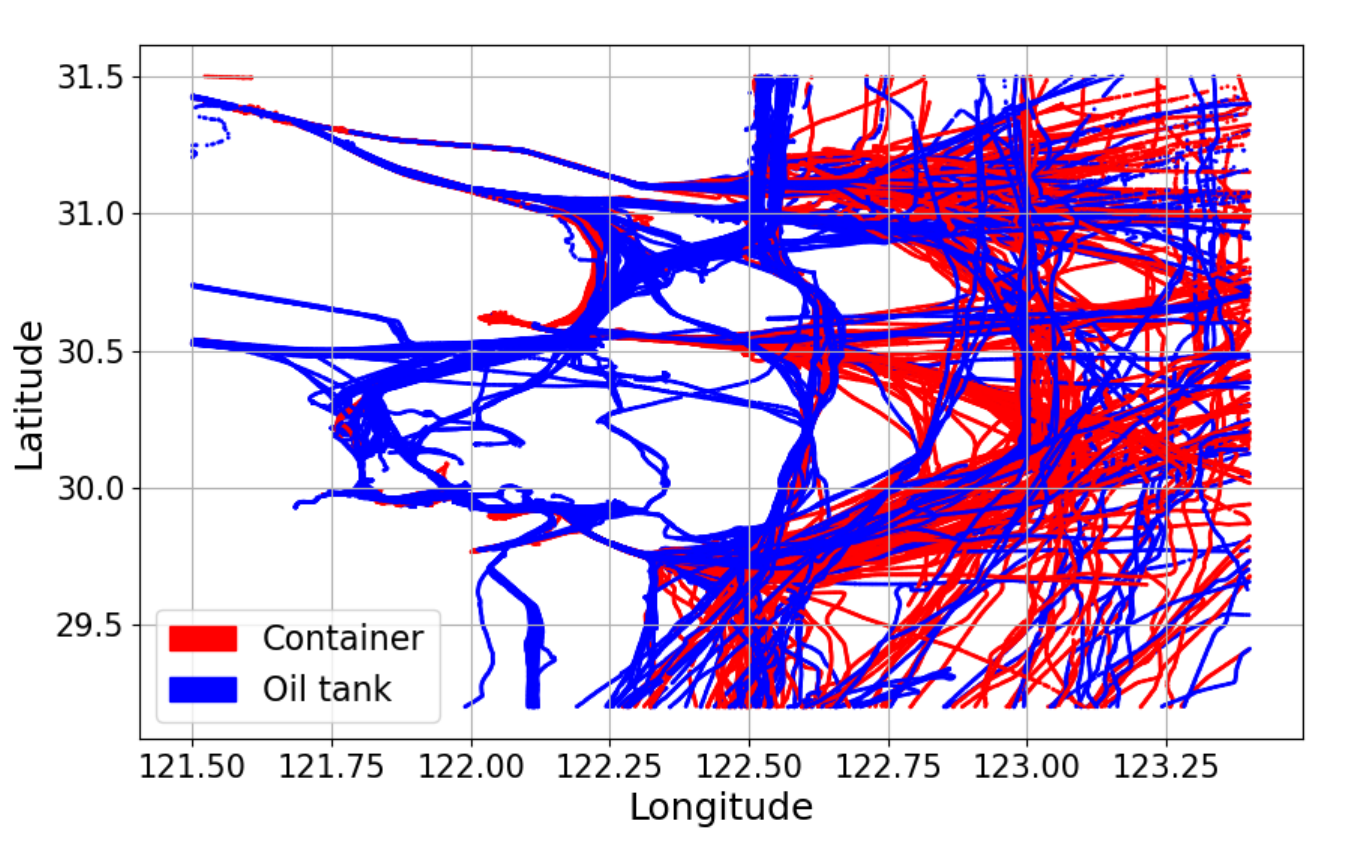}}
    \quad
    \subfloat[Clustering results]{
    \includegraphics[width=2.4in]{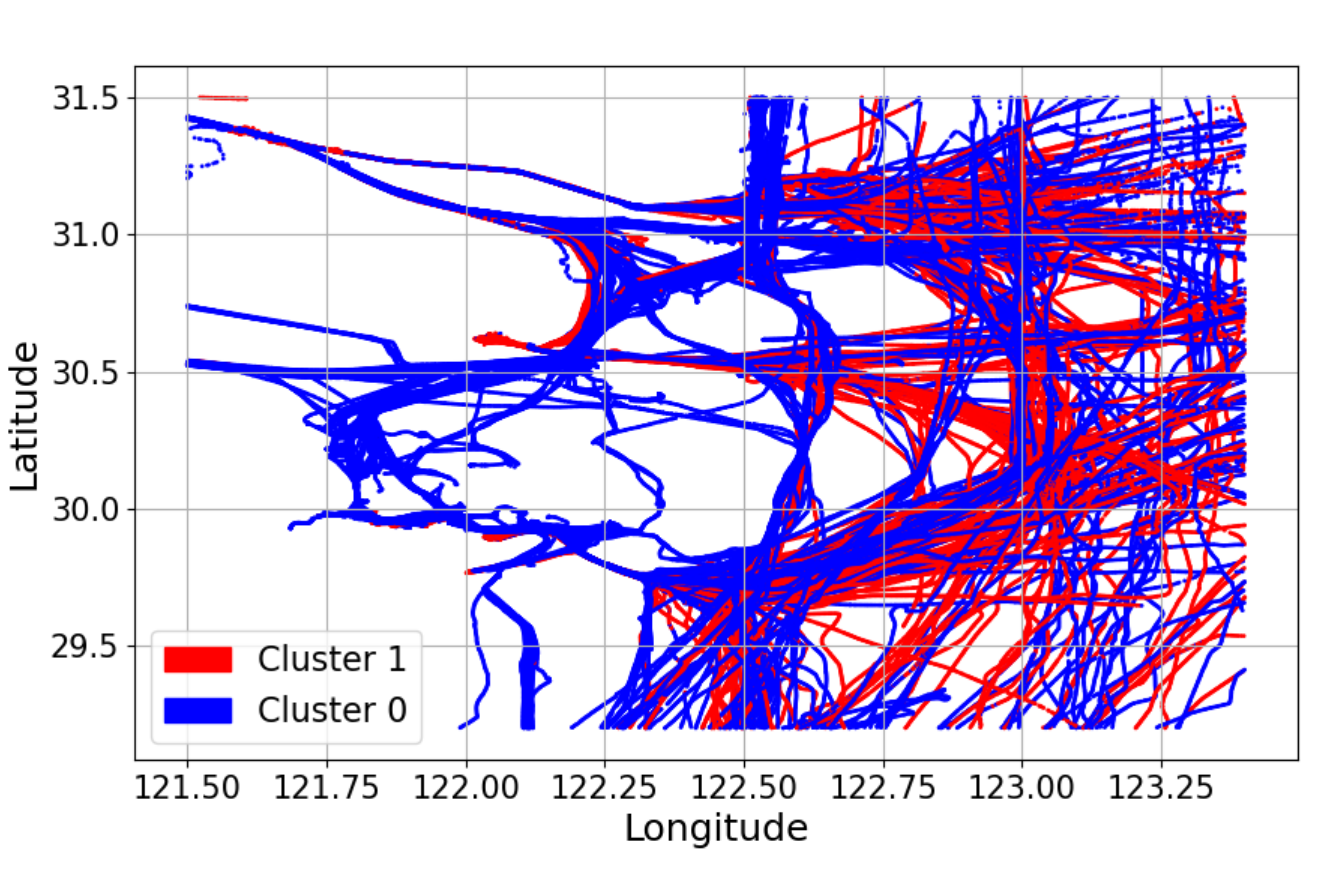}}
    \quad
    \caption{Clustering result of container vessel and oil tanker }
    \label{}
\end{figure*}

\begin{figure*}
    \centering
    \subfloat[containers]{
    \includegraphics[width=1.8in]{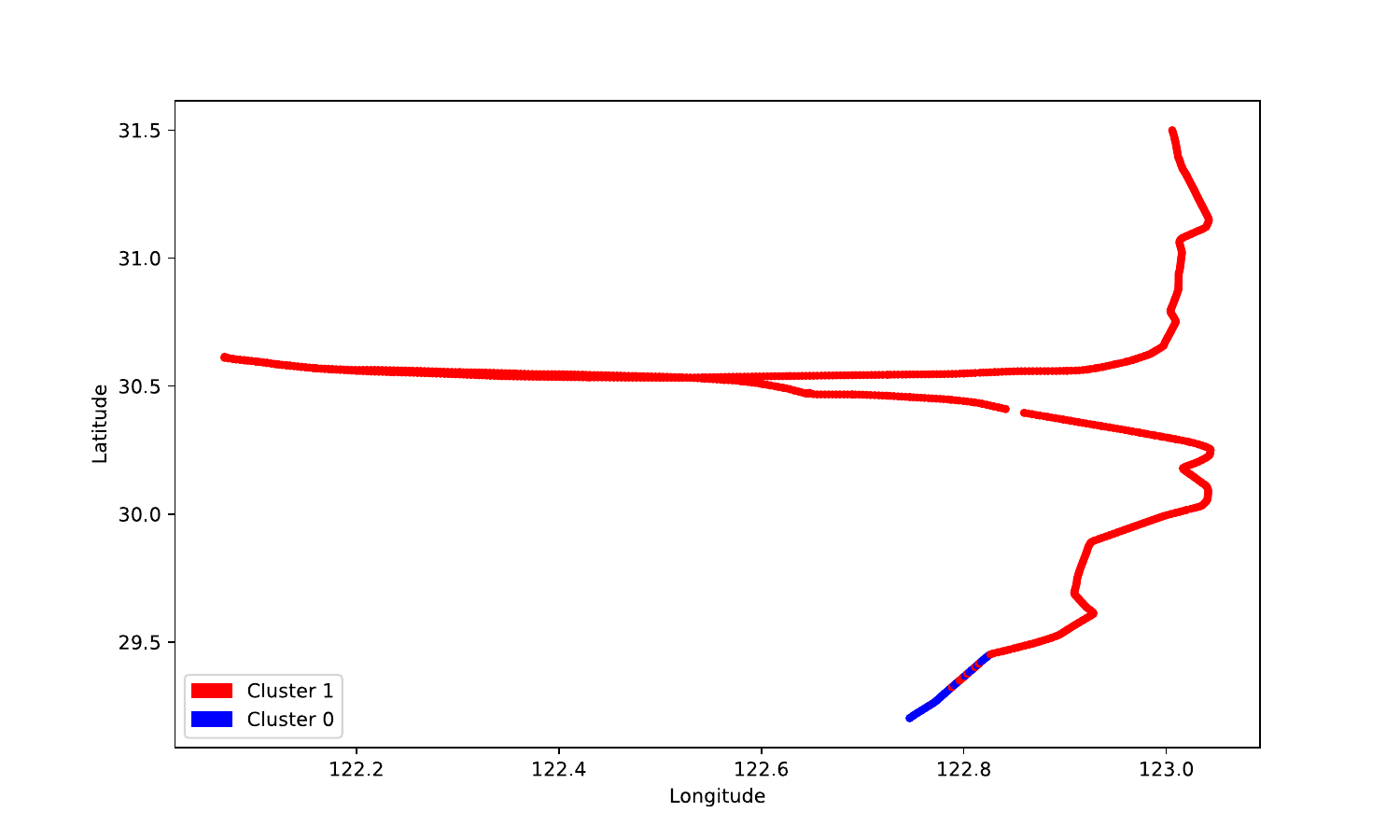}}
    \quad
    \subfloat[containers]{
    \includegraphics[width=1.8in]{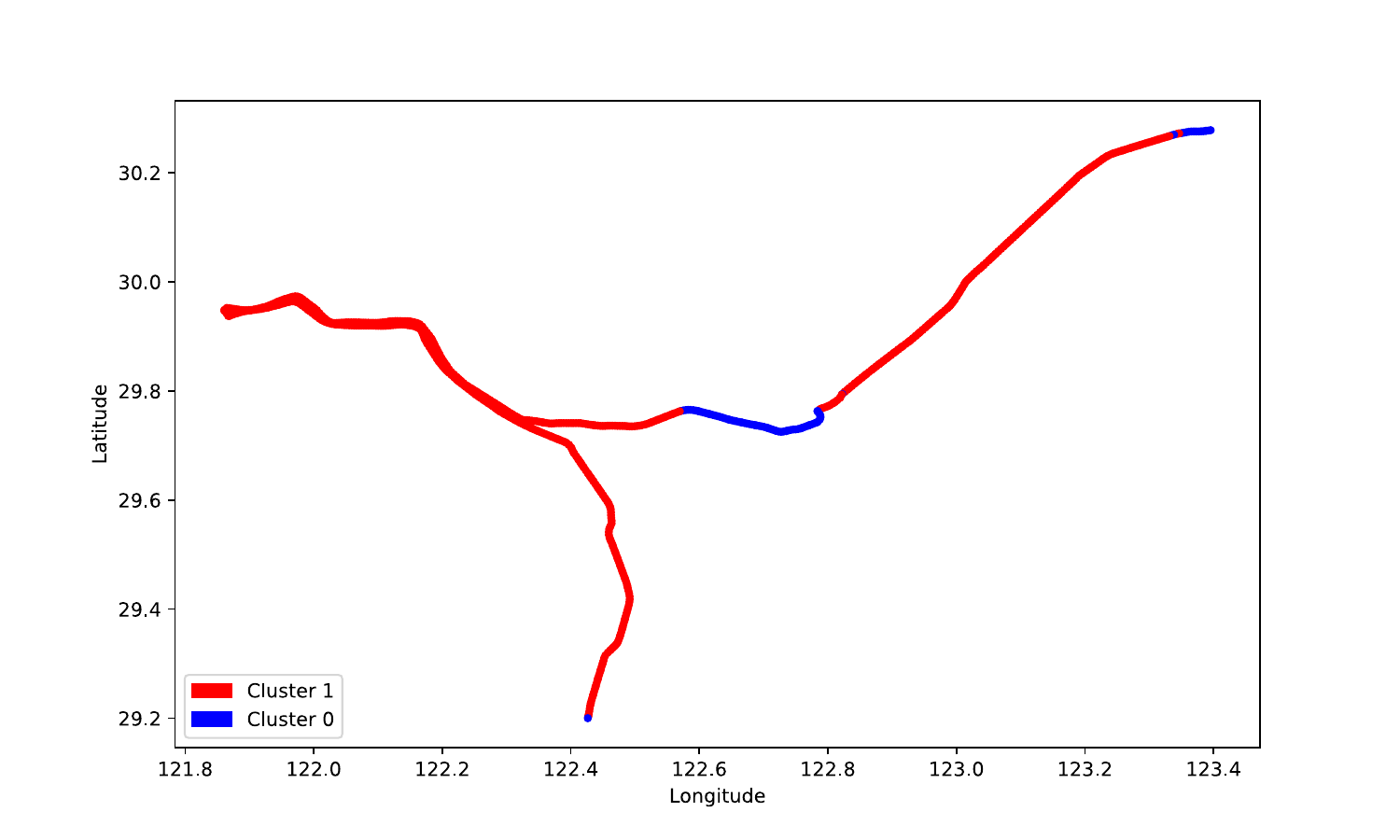}}
    \quad
    \subfloat[oil tanker]{
    \includegraphics[width=1.8in]{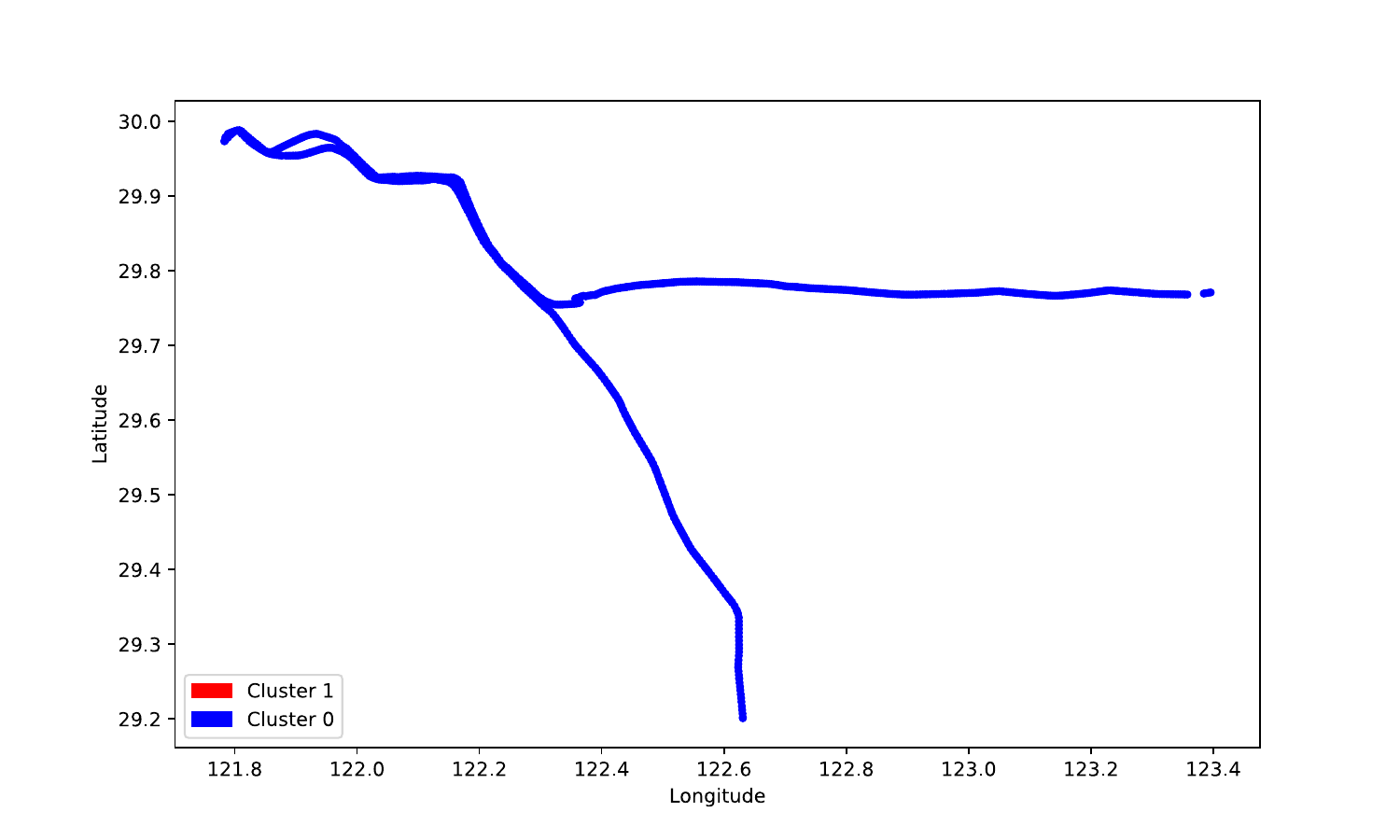}}
    \caption{Single trajectory analysis}
    \label{}
\end{figure*}
Figure 7 presents the results of the clustering for three different vessels. For the model to be accurate, the predicted label distributions for each group should be consistent. This means that each group has two distinct characteristics. \textit{i})  vessels of the same type should have similar types and times at every stage of their journey; \textit{ii}) the assigned type for each vessel should be updated over time. As vessels progress on their voyage, our model assigns them to either Cluster 0 or Cluster 1, and can quickly determine their vessel type.

The vessel shown in (a) is initially classified as part of Cluster 0. However, as its journey moves on, our model recognizes that it actually belongs to Cluster 1 and could accurately identify its type early on. The vessel displays in (b) is classified as part of Cluster 1 at the beginning. Note that some segments of its trajectory are classified as part of Cluster 0. However, as the vessel continues on its voyage, our model confirms that it belongs to Cluster 1. The vessel represented in (c) is easily determined and belongs to Cluster 0 based on its trajectory input alone. This demonstrates that PC-HiV can dynamically update the vessel's cluster assignments over time and promptly identify its type based on its significant maneuvering behavior.

Based on the locations where different types of vessels are moored, there are evident differences. The result suggests that hierarchical vessel trajectory representation has a significant effect on vessel behavior clustering. Different types of vessels tend to moor at specific geographic coordinates, and there are variations in their mooring locations.

\subsubsection{Case of In/Out ECA}
We analyze vessel trajectories at the ECA boundary of the emission control area. Emission Control Areas (ECAs), regional marine environmental measures, have been established by the International Maritime Organization (IMO). The Hainan dataset was analyzed to study vessel trajectories passing through ECAs. The analysis involved sub-trajectory sequence, and the results revealed that vessels tend to slow down or anchor near the ECA boundaries. 
\begin{figure*}[h] 
  \centering
  \begin{minipage}{0.33\textwidth}
    \centering
    \includegraphics[width=1\linewidth]{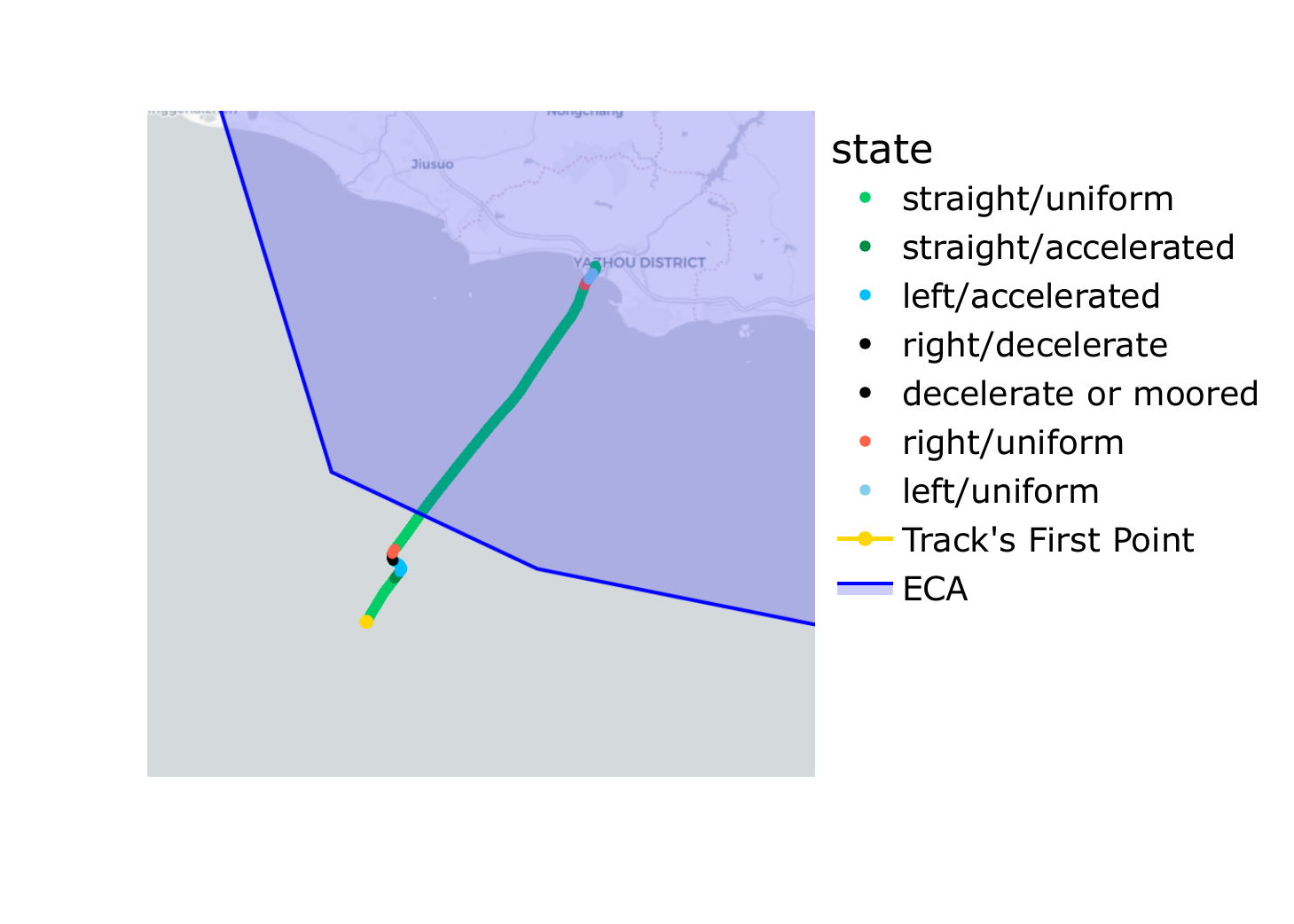}
  \end{minipage}%
  \begin{minipage}{0.33\textwidth}
    \centering
    \includegraphics[width=1\linewidth]{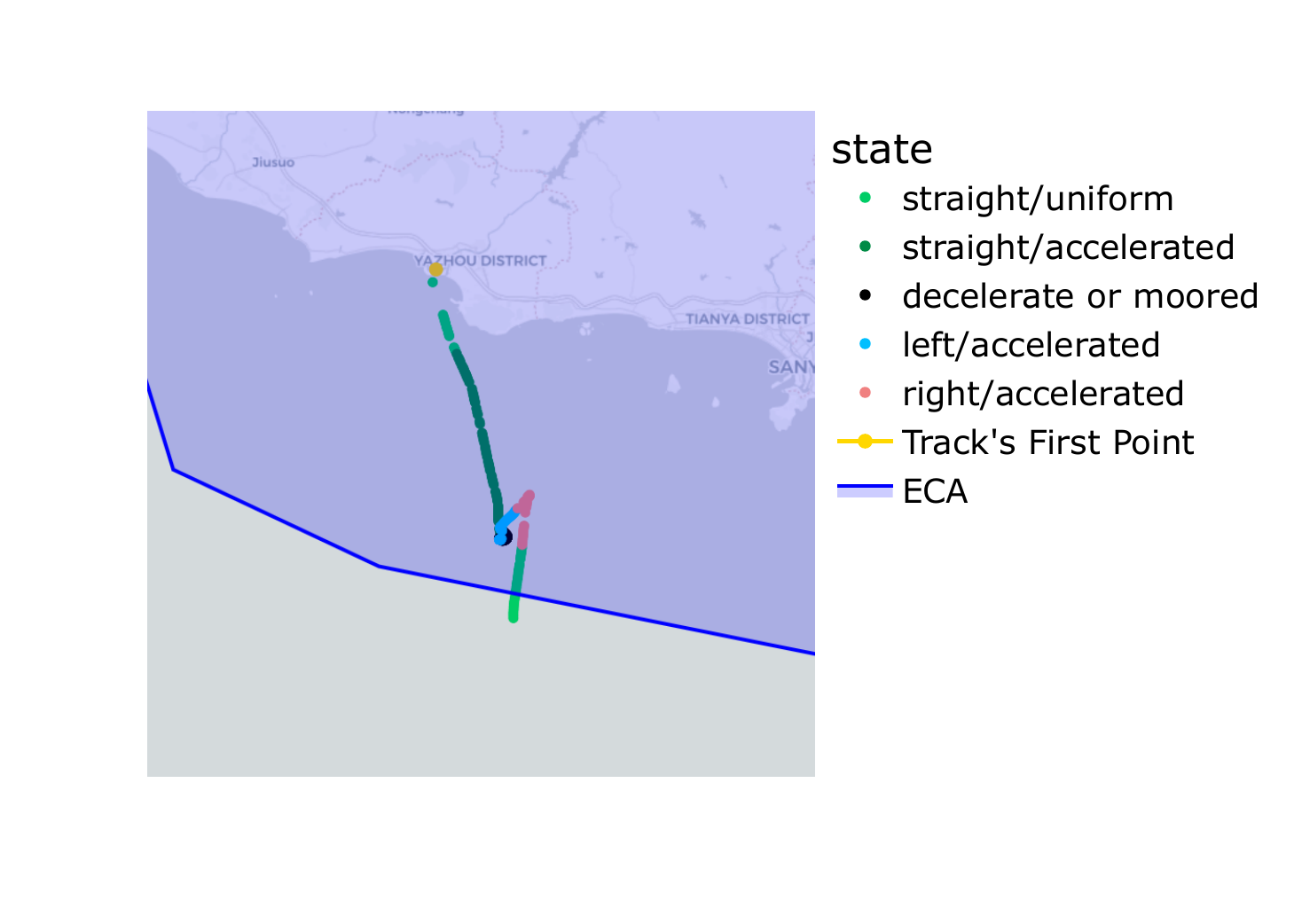}
  \end{minipage}%
  \begin{minipage}{0.33\textwidth}
    \centering
    \includegraphics[width=1\linewidth]{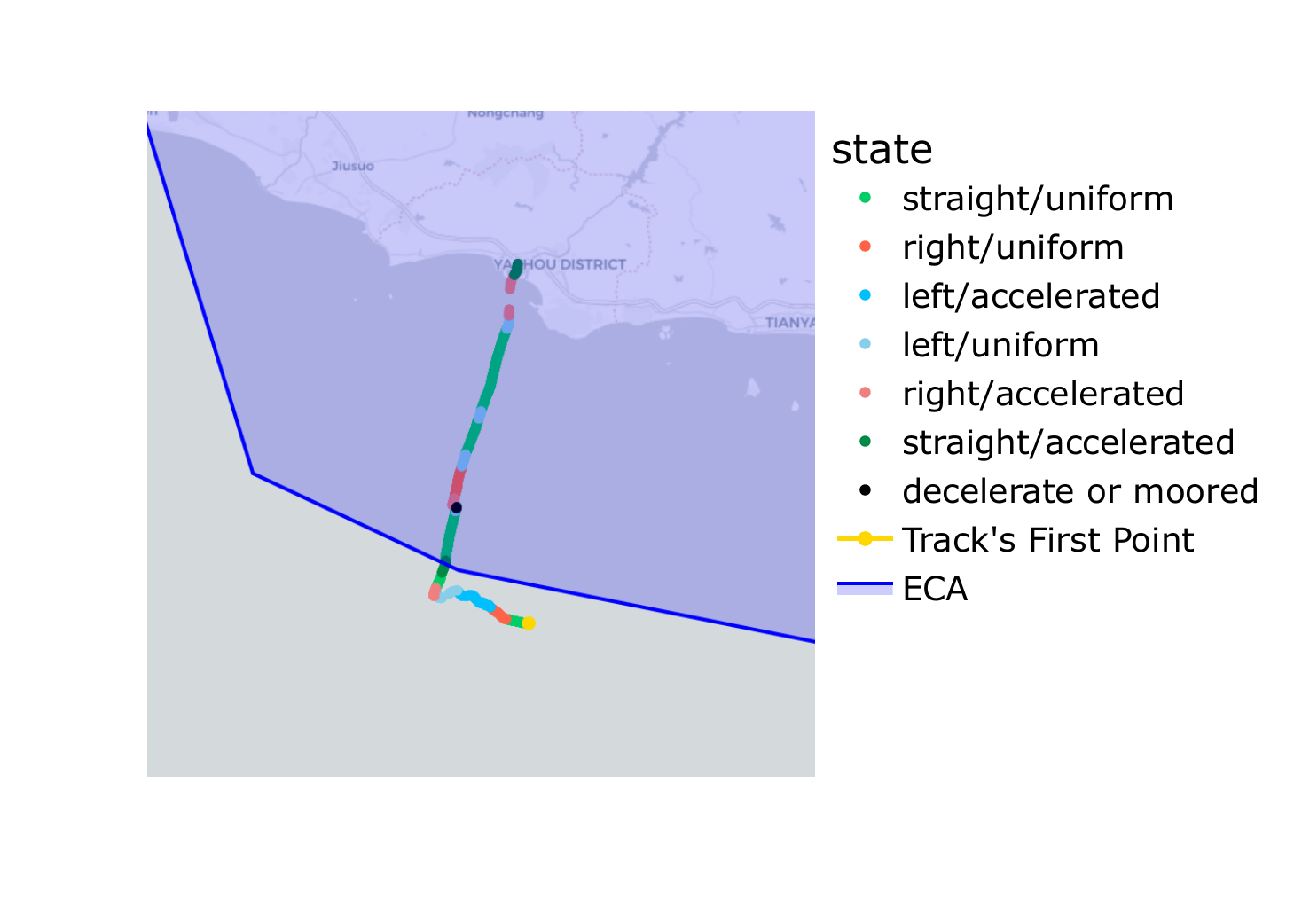}
  \end{minipage}%

  \begin{minipage}{0.33\textwidth}
    \centering
    \includegraphics[width=1\linewidth]{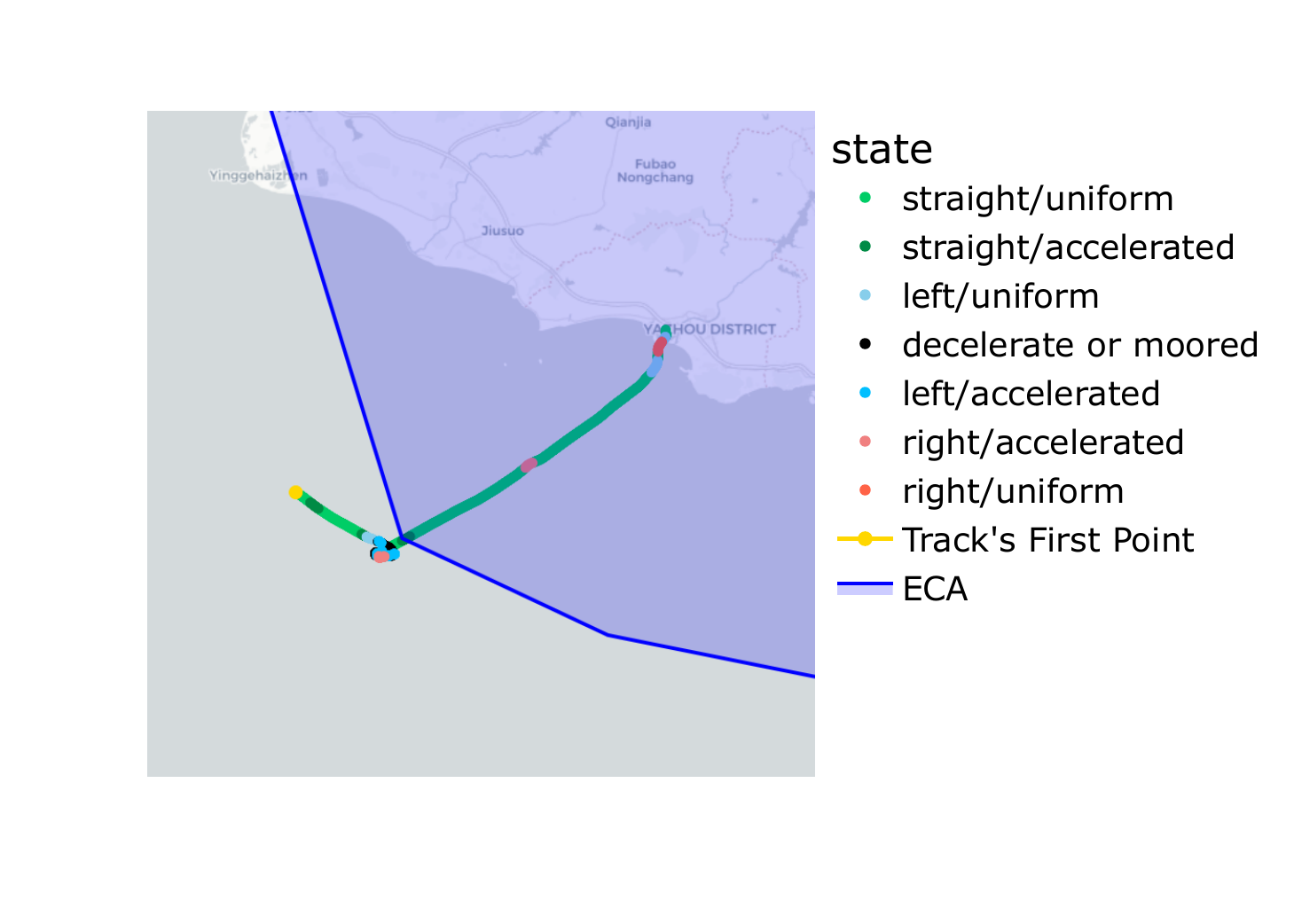}
  \end{minipage}
  \begin{minipage}{0.33\textwidth}
    \centering
    \includegraphics[width=1\linewidth]{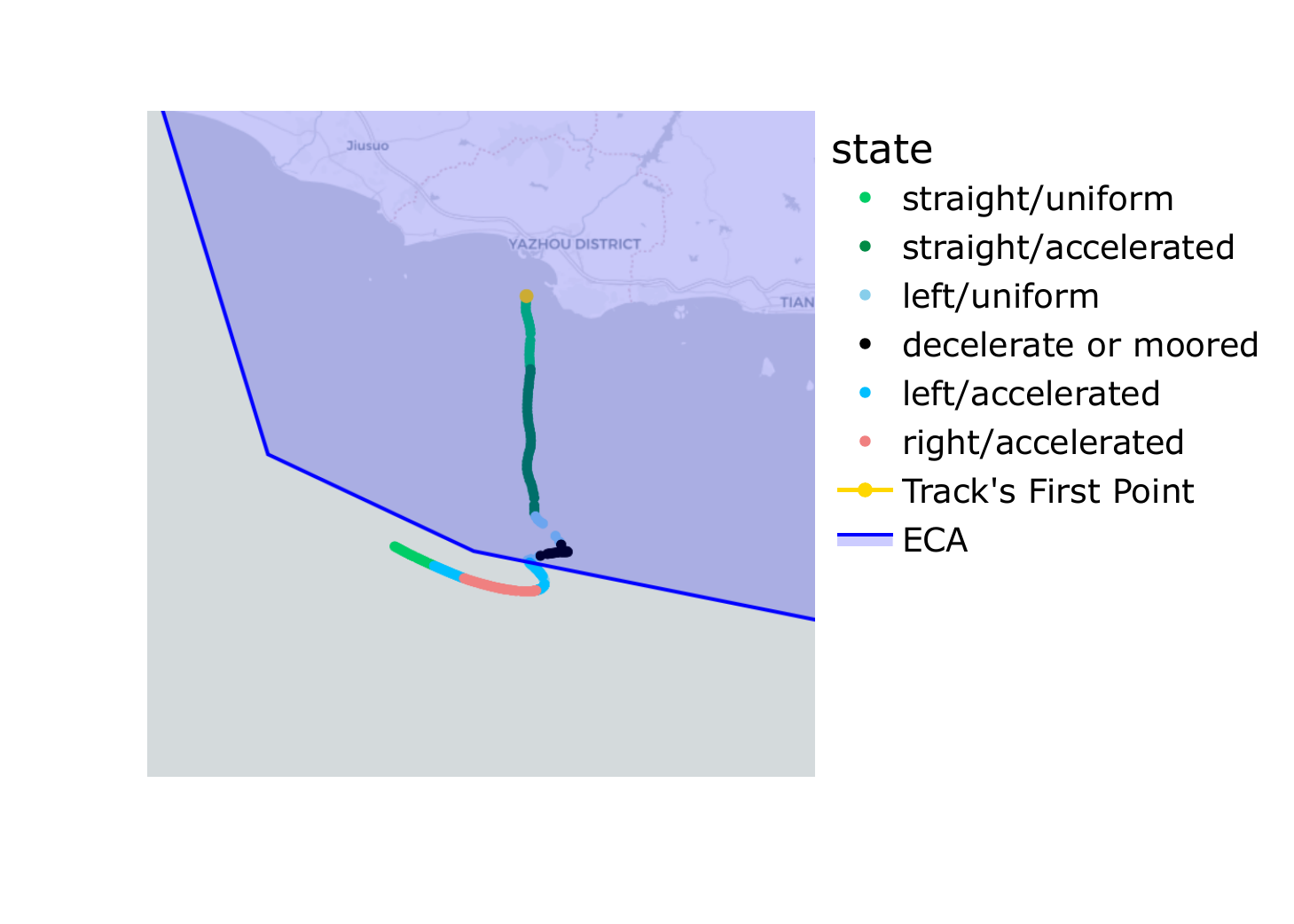}
  \end{minipage}%
  \begin{minipage}{0.33\textwidth}
    \centering
    \includegraphics[width=1\linewidth]{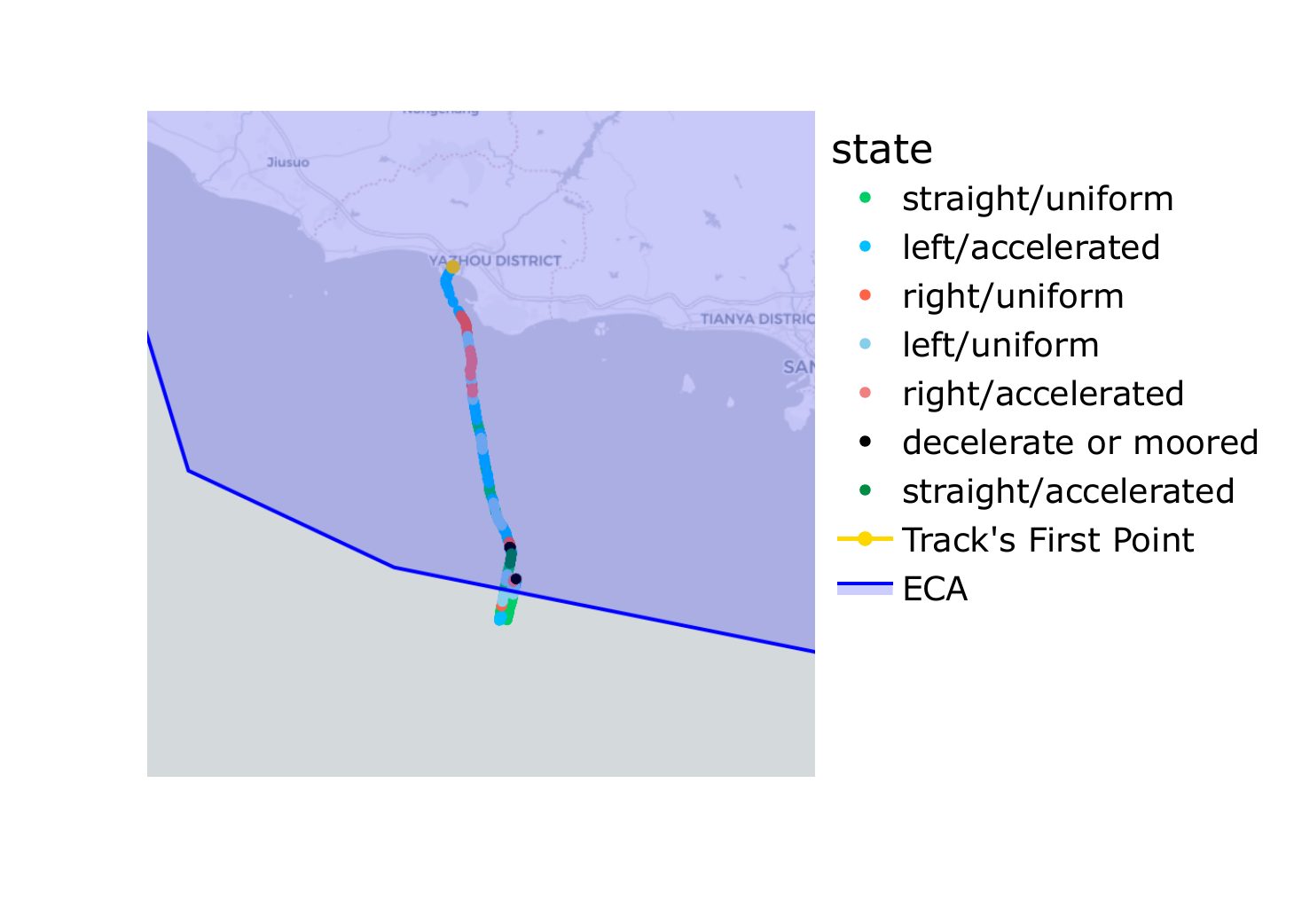}
  \end{minipage}%
  \caption{Vessel trajectories at the ECA boundary}
  \label{fig:grid}
\end{figure*}

According to Figure 8, vessels that go through the controlled area and slow down significantly as they approach the Emission Control Area (ECA). They tend to change direction near the ECA boundary. Interestingly, the vessels that decelerate do not continue forward but opt to change direction instead. Based on the path they take after changing course, it seems that this change in direction does not help the vessels to reach their intended destinations. Instead, they move parallel to the ECA boundary while  slowing down. This behavior  suggests that the vessels are considering changing their fuel before entering or exiting the ECA.

\subsection{Limitation analysis}
Our research is centered on vessel traffic in ports and waterways nearby. Although our main focus is on port-related trajectories, our model has the potential to handle data from a wider geographical range. However, it's important to note that there are limitations to expanding the analysis to a larger area.
\begin{enumerate}
\item{\textbf{\textbf{Design and application of label sequences}. } The limitation of our study pertains to the design of the mapping function from sub-trajectory sequence to label sequence. Incorporating domain-specific knowledge, such as local navigational regulations and practices, could lead to more effective label sequences. However, obtaining and integrating such domain knowledge can be challenging. In Specific application scenarios, such as "Regulated traffic areas" (RTAs), pollution monitoring, or fisheries management, may require distinct label sequences to account for varying aspects of vessel behavior. Thus, the generalizability of our label sequence design across different maritime applications remains a limitation.}
\item{\textbf{Hydrological features.} Our model does not take hydrological features into account. Hydrological features encompass factors such as tides, currents, water depth, and more, which have a significant impact on vessel behavior and navigation decisions.}
\end{enumerate}

\section{Conclusion}\label{sec:conclusion}

In this paper, we propose the so-called PC-HiV to cluster vessel behaviors in hierarchies. Hierarchical representations can provide more precise feature descriptions of trajectories. The clustering results based on them show that they can learn better representations compared to traditional trajectory representation methods. Besides, PC-HiV uses predictive clustering to improve the results, as it can update cluster and behavior predictions simultaneously.

In the future, our method can be expanded from two aspects. On the one hand, the method to select port labels in the hierarchical representations requires further optimization. At the current stage, our method is rather brute-force. We hope that attribute information on mooring positions can help the process of selection. On the other hand, in a practical situation where two or more vessels will meet, it is not sufficient to only use current behavior labels. Hence, it is worth investigating using PC-HiV to represent meeting behaviors and  avoid the collision.

\end{document}